\documentclass[10pt,twocolumn,letterpaper]{article}

\usepackage{cvpr}
\usepackage{times}
\usepackage{epsfig}
\usepackage{graphicx}
\usepackage{amsmath}
\usepackage{amssymb}
\usepackage{tabu}
\usepackage{float}
\usepackage{subfigure}
\usepackage{multirow}
\usepackage{array}
\usepackage{enumerate}
\usepackage{booktabs}
\usepackage{epstopdf}

\usepackage[breaklinks=true,bookmarks=false]{hyperref}

\makeatletter
\renewcommand{\@fnsymbol}[1]{\ensuremath{%
   \ifcase#1\or \# \or *\or {*}*\or
   \mathsection\or \mathparagraph\or \|\or \star\or
   \star\star\or {\star\star}\star \else\@ctrerr\fi}}
\makeatother

\cvprfinalcopy % *** Uncomment this line for the final submission

\def\cvprPaperID{1989} % *** Enter the CVPR Paper ID here

\begin{document}

%%%%%%%%% TITLE
\title{Exploiting Kernel Sparsity and Entropy for Interpretable CNN Compression}

\author{
    Yuchao Li$^{1\#}$,
    Shaohui Lin$^{1}$\thanks{Equal contribution.} \ ,
    Baochang Zhang$^{3}$,
    Jianzhuang Liu$^{4}$, \\
    David Doermann$^{5}$,
    Yongjian Wu$^{6}$,
    Feiyue Huang$^{6}$,
    Rongrong Ji$^{1,2}$\thanks{Corresponding author.}
    \\
    $^{1}$Fujian Key Laboratory of Sensing and Computing for Smart City, Department of Cognitive Science, \\
    School of Information Science and Engineering, Xiamen University, Xiamen, China\\
    $^{2}$Peng Cheng Laboratory, Shenzhen, China,
    $^{3}$Beihang University, China,
    $^{4}$Huawei Noah’s Ark Lab\\
    $^{5}$University at Buffalo, USA,
    $^{6}$BestImage, Tencent Technology (Shanghai) Co.,Ltd, China
    \\
    {\tt\small xiamenlyc@gmail.com,}
    {\tt\small shaohuilin007@gmail.com,}
    {\tt\small bczhang@buaa.edu.cn,}
    {\tt\small liu.jianzhuang@huawei.com,} \\
    {\tt\small doermann@buffalo.edu,}
    {\tt\small littlekenwu@tencent.com,}
    {\tt\small garyhuang@tencent.com,}
    {\tt\small rrji@xmu.edu.cn}
    % \IEEEauthorblockN{Yuchao Li}
    % \IEEEauthorblockA{Xiamen University\\
    % xiamenlyc@stu.xmu.edu.cn}\\
    % % \and
    % \IEEEauthorblockN{Xiaopeng Hong}
    % \IEEEauthorblockA{University of Oulu, Finland\\
    % xiaopeng.hong@oulu.fi}
    % \and
    % \IEEEauthorblockN{Rongrong Ji}
    % \IEEEauthorblockA{Xiamen University\\
    % rrji@xmu.edu.cn}\\
    % % \and
    % \IEEEauthorblockN{Yue Gao}
    % \IEEEauthorblockA{Tsinghua University\\
    % kevin.gaoy@gmail.com}
    % \and
    % \IEEEauthorblockN{Hong Liu}
    % \IEEEauthorblockA{Xiamen University\\
    % lynnliu.xmu@gmail.com}\\
    % % \and
    % \IEEEauthorblockN{Qi Tian}
    % \IEEEauthorblockA{Huawei Noah's Ark Lab\\
    % tian.qi1@huawei.com}
}

\maketitle
%\thispagestyle{empty}

%%%%%%%%% ABSTRACT
\begin{abstract}
  Compressing convolutional neural networks (CNNs) has received ever-increasing research focus.
  However, most existing CNN compression methods do not interpret their inherent structures to distinguish the implicit redundancy.
  In this paper, we investigate the problem of CNN compression from a novel interpretable perspective.
  The relationship between the input feature maps and $2$D kernels is revealed in a theoretical framework, based on which a kernel sparsity and entropy (KSE) indicator is proposed to quantitate the feature map importance in a \textit{feature-agnostic} manner to guide model compression.
  Kernel clustering is further conducted based on the KSE indicator to accomplish high-precision CNN compression.
  KSE is capable of simultaneously compressing each layer in an efficient way, which is significantly faster compared to previous data-driven feature map pruning methods.
  We comprehensively evaluate the compression and speedup of the proposed method on CIFAR-10, SVHN and ImageNet 2012.
  Our method demonstrates superior performance gains over previous ones.
  In particular, it achieves $4.7 \times$ FLOPs reduction and $2.9 \times$ compression on ResNet-50 with only a Top-5 accuracy drop of $0.35\%$ on ImageNet 2012, which significantly outperforms state-of-the-art methods.

\end{abstract}

%%%%%%%%% BODY TEXT
\section{Introduction}

\begin{figure}[t]
\begin{center}
  \includegraphics[width=0.4\textwidth]{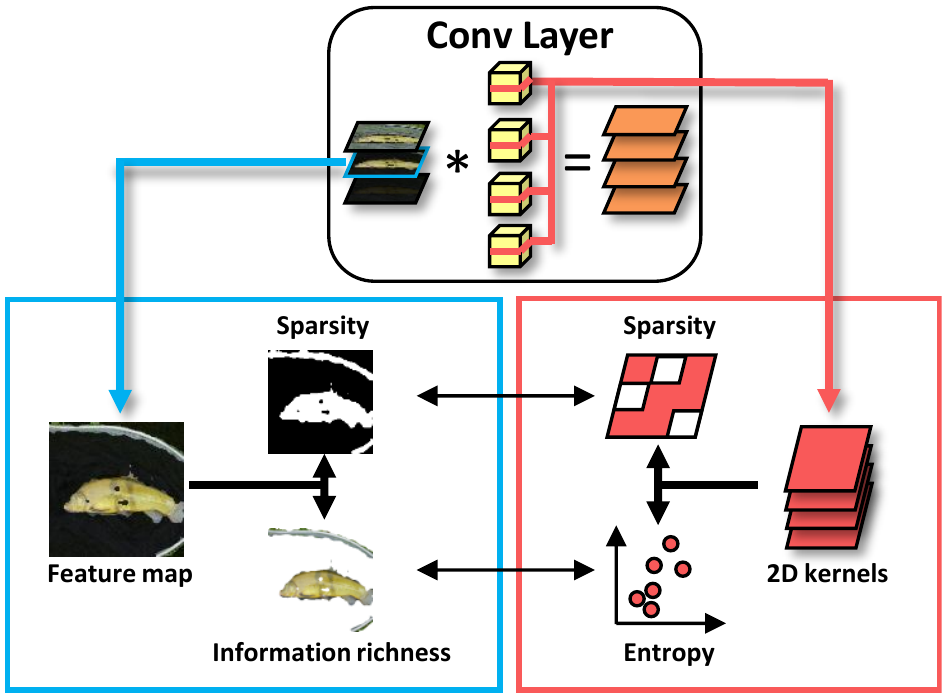}
\end{center}
\vspace{-.5em}
   \caption{The relationship between an input feature map and its corresponding $2$D kernels is investigated. We introduce the kernel sparsity and entropy (KSE) to represent the sparsity and information richness of the input feature maps.}
   \vspace{-1.5em}
\label{fig:first}
\end{figure}

\begin{figure*}
  \setlength\abovecaptionskip{-1em}
\begin{center}
\vspace{-1em}
\includegraphics[width=0.97\linewidth]{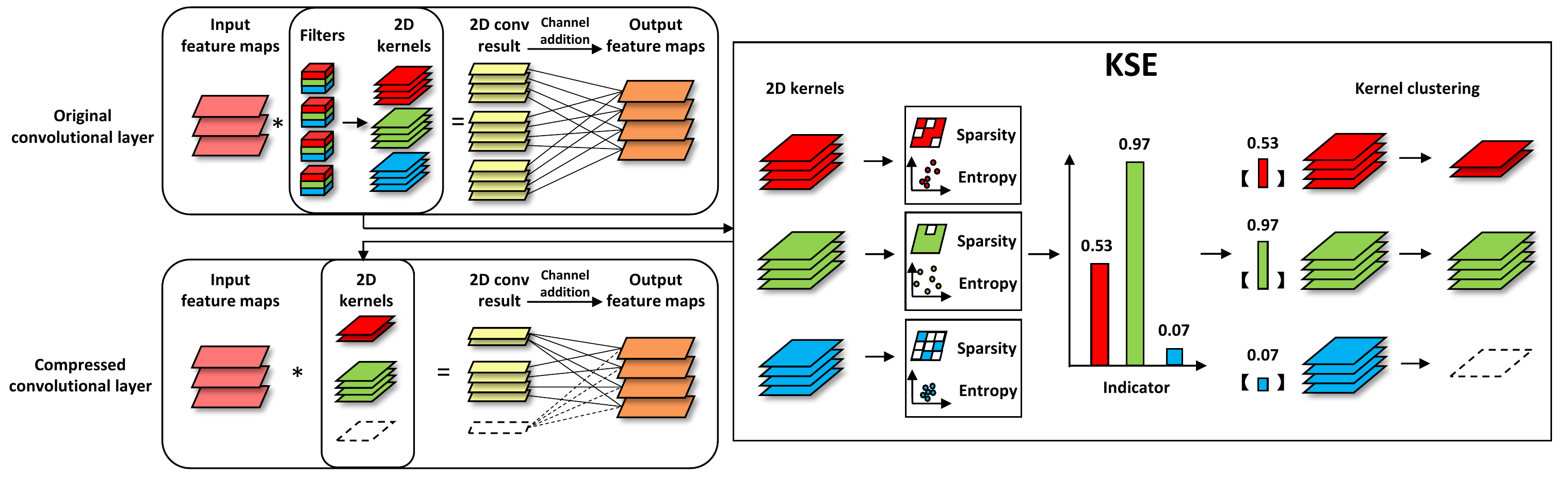}
\end{center}
\vspace{-1em}
   \caption{The framework of our method. The convolution operation is split into two parts, $2$D convolution and channel fusion (addition). The $2$D convolution is used to extract features from each input feature map, and the channel addition is used to obtain an output feature map by summing the intermediate results of the $2$D convolution across all the input feature maps. In our KSE method, we first obtain the $2$D kernels corresponding to an input feature map and calculate their sparsity and entropy as an indicator, which is further used to reduce the number of the $2$D kernels by kernel clustering and generate a compact network.}
   \vspace{-1em}
\label{fig:short}
\end{figure*}

Deep convolutional neural networks (CNNs) have achieved great success in various computer vision tasks, including object classification \cite{he2016deep,huang2017densely,krizhevsky2012imagenet,simonyan2014very}, detection \cite{redmon2017yolo9000,Ren2015Faster} and semantic segmentation \cite{chen2018encoder,Long2015Fully}.
However, deep CNNs typically require high computation overhead and large memory footprint, which prevents them from being directly applied on mobile or embedded devices.
As a result, extensive efforts have been made for CNN compression and acceleration, including low-rank approximation \cite{lebedev2014speeding, lin2018holistic, lin2016towards, wen2017coordinating}, parameter quantization \cite{jacob2017quantization, zhou2016dorefa} and binarization \cite{rastegari2016xnor}.
One promising direction to reduce the redundancy of CNNs is network pruning \cite{han2015deep, han2015learning, hu2016network, huang2018data, lin2019towards, luo2017thinet, yoon2017combined}, which can be applied to different elements of CNNs such as the weights, the filters and the layers.

Early works in network pruning \cite{han2015deep, han2015learning} mainly resort to removing less important weight connections independently with precision loss as little as possible.
However, these unstructured pruning methods require specialized software or hardware designs to store a large number of indices for efficient speedup.
Among them, filter pruning has received ever-increasing research attention, which can simultaneously reduce computation complexity and memory overhead by directly pruning redundant filters, and is well supported by various off-the-shelf deep learning platforms.
For instance, Molchanov \emph{et al}. \cite{molchanov2016pruning} calculated the effect of filters on the network loss based on a Taylor expansion.
Luo \emph{et al.} \cite{luo2017thinet} proposed to remove redundant filters based on a greedy channel selection.
Those methods directly prune filters and their corresponding output feature maps in each convolutional layer, which may lead to dimensional mismatch in popular multi-branch networks, \emph{e.g.,}\ ResNets \cite{he2016deep}.
% and ResNeXts \cite{xie2017aggregated}.
%
For example, by removing the output feature maps in the residual mapping, the ``add'' operator cannot be implemented due to different output dimensions between the identity mapping and the residual mapping in ResNets.
Instead, several channel pruning methods \cite{huang2018condensenet, liu2017learning} focus on the input feature maps in the convolutional layers, which do not modify the network architecture and operator when reducing the network size and FLOPs\footnote{FLOPs: The number of floating-point operations}.
However, through directly removing the input feature maps, this approach typically has limited compression and speedup with significant accuracy drop.

In this paper, we investigate the problem of CNN compression from a novel \textit{interpretable} perspective.
We argue that interpreting the inherent network structure provides a novel and fundamental means to discover the implicit network redundancy.
As investigated in network explanation \cite{li2015convergent,morcos2018importance,zhou2018revisiting}, individual feature maps within and across different layers play different roles in the network.
As an intuitive example, feature maps in different layers can be seen as hierarchical features, \emph{e.g.,}\ features like simple structures in the bottom layers, and semantic features in the top layers.
Even in the same layer, the importance of feature maps varies; the more information a feature map represents, the more important it is for the network.
To this end, interpreting the network, especially the feature map importance, if possible, can well guide the quantization and/or pruning of the network elements.

We here have the first attempt to interpret the network structure towards fast and robust CNN compression.
In particular, we first introduce the receptive field of a feature map to reveal the sparsity and information richness, which are the key elements to evaluate the feature map importance.
Then, as shown in Fig. \ref{fig:first}, the relationship between an input feature map and its corresponding 2D kernels is investigated, based on which we propose kernel sparsity and entropy (KSE) as a new indicator to efficiently quantitate the importance of the input feature maps in a \textit{feature-agnostic} manner.
Compared to previous data-driven compression methods \cite{he2017channel, hu2016network, luo2017thinet}, which need to compute all the feature maps corresponding to the entire training dataset to achieve a generalized result, and thus suffer from heavy computational cost for a large dataset, KSE can efficiently handle every layer in parallel in a data-free manner.
Finally, we employ kernel clustering to quantize the kernels for CNN compression, and fine-tune the network with a small number of epochs.

We demonstrate the advantages of KSE using two widely-used models (ResNets and DenseNets) on three datasets (CIFAR-10, SVHN and ImageNet 2012).
Compared to the state-of-the-art methods, KSE achieves superior performance.
For ResNet-50, we obtain $4.7 \times$ FLOPs reduction and $2.9 \times$ compression with only $0.35\%$ Top-5 accuracy drop on ImageNet.
The compressed DenseNets achieve much better performance than other compact networks (\emph{e.g.,}\ MobileNet V2 and ShuffleNet V2) and auto-searched networks (\emph{e.g.,}\ MNASNet and PNASNet).

The main contributions of our paper are three-fold:
\begin{itemize}
\vspace{-.5em}
\item We investigate the problem of CNN compression from a novel \textit{interpretable} perspective, and discover that the importance of a feature map depends on its sparsity and information richness.
\vspace{-.7em}
\item Our method in Fig. \ref{fig:short} is \textit{feature-agnostic} that only needs the $2$D kernels to calculate the importance of the input feature maps, which differs from the existing data-driven methods based on directly evaluating the feature maps \cite{he2017channel, hu2016network, luo2017thinet}.
It can thus simultaneously handle all the layers efficiently in parallel.
\vspace{-.7em}
\item Kernel clustering is proposed to replace the common kernel pruning methods \cite{mao2017exploring, wen2016learning}, which leads to a higher compression ratio with only slight accuracy degradation.
\end{itemize}
%The rest of this paper is organized as follows: In Section \ref{sec:2}, we review the related work. Then we interpret the feature map importance in Section \ref{sec:3} and describe our KSE method in Section \ref{sec:4}. In Section \ref{sec:5}, we report the quantitative results of KSE. Finally, we conclude this paper in Section \ref{sec:6}.

%------------------------------------------------------------------------

\section{Related Work}
\label{sec:2}

In this section, we briefly review related work of network pruning for CNN compression that removes redundant parts, which can be divided into unstructured pruning and structured pruning.

Unstructured pruning is to remove unimportant weights independently.
Han \emph{et al}. \cite{han2015deep, han2015learning} proposed to prune the weights with small absolute values, and store the sparse structure in a compressed sparse row or column format.
Yang \emph{et al.} \cite{yang2016designing} proposed an energy-aware pruning approach to prune the unimportant weights layer-by-layer by minimizing the error reconstruction.
Unfortunately, these methods need a special format to store the network and the speedup can only be achieved by using specific sparse matrix multiplication in the special software or hardware.

By contrast, structured pruning directly removes structured parts (\emph{e.g.,}\ kernels, filters or layers) to simultaneously compress and speedup CNNs and is well supported by various off-the-shelf deep learning libraries.
Li \emph{et al}. \cite{li2016pruning} proposed to remove unimportant filters based on the $\ell_1$-norm.
Hu \emph{et al}. \cite{hu2016network} computed the Average Percentage of Zeros (APoZ) of each filter, which is equal to the percentage of zero values in the output feature map corresponding to the filter.
Recently, Yoon \emph{et al}. \cite{yoon2017combined} proposed a group sparsity regularization that exploits correlations among features in the network.
He \emph{et al}. \cite{he2017channel} proposed a LASSO regression based channel selection, which uses least square reconstruction to prune filters.
Differently, Lin \emph{et al}. \cite{lin2018accelerating} proposed a global and dynamic training algorithm to prune unsalient filters.
Although filter pruning approaches in \cite{he2017channel,hu2016network,lin2018accelerating,luo2017thinet,yoon2017combined} can reduce the memory footprint, they encounter a dimensional mismatch problem for the popular multi-branch networks, \emph{e.g.,}\ ResNets \cite{he2016deep}.
Our method differs from all the above ones, which reduces the redundancy of $2$D kernels corresponding to the input feature maps and do not modify the output of a convolutional layer to avoid the dimensional mismatch.

Several channel pruning methods \cite{huang2018condensenet,liu2017learning} are more suitable for the widely-used ResNets and DenseNets.
These methods remove unimportant input feature maps of convolutional layers, and can avoid dimensional mismatch.
For example, Liu \emph{et al}. \cite{liu2017learning} imposed $\ell_1$-regularization on the scaling factors on the batch normalization to select unimportant feature maps.
Huang \emph{et al}. \cite{huang2018condensenet} combined the weight pruning and group convolution to sparsify networks.
These channel pruning methods obtain a sparse network based on a
complex training procedure that requires significant cost of offline training.
%
% Unlike these methods, our approach reduces the 2D convolutional kernels that extract features on the input feature maps in a \textit{feature-agnostic} manner, instead of directly removing them, which is able to achieve finer-grained compression more faster.
Unlike these methods, our approach determines the importance of the feature maps in a novel interpretable perspective and calculates them by their corresponding kernels without extra training, which is significantly faster to implement CNNs compression.

Besides the network pruning, our work is also related to some other methods \cite{son2018clustering, tung2018clip,wu2016quantized}.
Wu \emph{et al}. \cite{wu2016quantized} quantified filters in convolutional layers and weight matrices in fully-connected layers by minimizing the reconstruction error.
However, they do not consider the different importance of input feature maps, and instead the same number of quantification centroids.
%
% Tung \emph{et al}. \cite{tung2018clip} combined weight pruning and quantization to compress CNN, which is an unstructured compression that needs specialized software and hardware designs.
Wu \emph{et al}. \cite{wu2018deep} applied k-means clustering on the weights to compress CNN, which needs to set the different number of cluster centroids.
Son \emph{et al}. \cite{son2018clustering} compressed the network by using a small amount of spatial convolutional kernels to minimize the reconstruction error.
However, it is mainly used for $3\times 3$ kernels and difficult to compress $1\times 1$ kernels.
In contrast, our KSE method can be applied to all layers.

Note that our approach can be further integrated with other strategies to obtain more compact networks, such as low-rank approximation \cite{lin2017espace, lin2018holistic, lin2016towards, wen2017coordinating} and compact architecture design \cite{ma2018shufflenet,sandler2018mobilenetv2}.

%------------------------------------------------------------------------
\section{Interpretation of Feature Maps}
\label{sec:3}

\begin{figure}[t]
\begin{center}
  \includegraphics[width=0.47\textwidth]{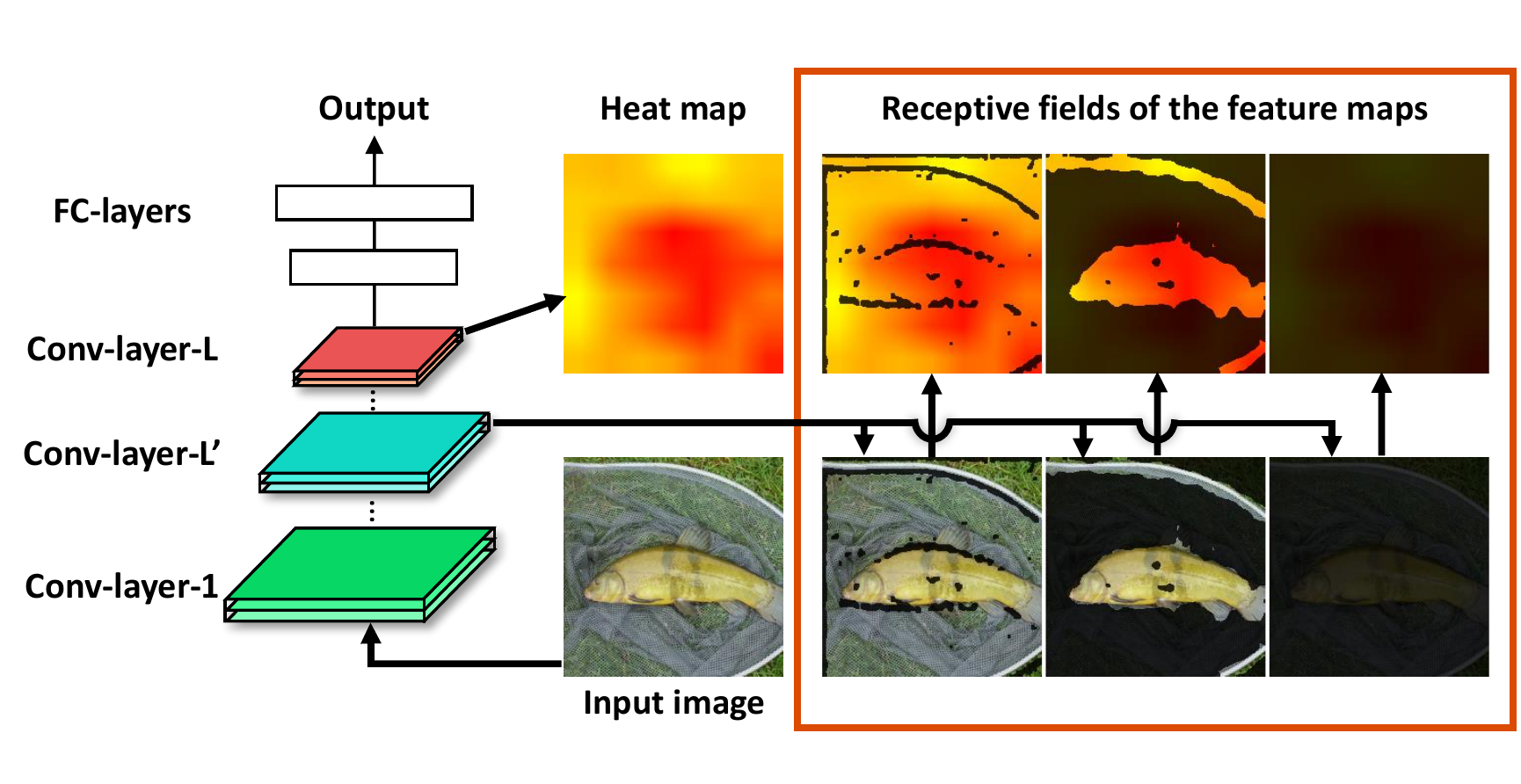}
\end{center}
\vspace{-1em}
   \caption{Visualization of the receptive fields (in the resolution of the input image) of three feature maps in the network.}
\vspace{-1em}
\label{fig:exp}
\end{figure}

Towards identifying feature map importance on the network, the work in \cite{bau2017network} uses bilinear interpolation to scale the feature maps up to the resolution of the input image.
Then, a threshold determined by a top quantile level is used to obtain the receptive field of the feature maps, which can be regarded as a binary mask.
Following this principle, as shown in Fig. \ref{fig:exp}, for the input feature maps from the same convolutional layer, we compute their corresponding receptive fields on the original input image.
These receptive fields indicate the different information contained in these feature maps.
The visualization results in the lower part of the red box in Fig. \ref{fig:exp} can be interpreted as an indicator of the feature map importance, where the left one is the most important among the three while the right one is unimportant.

To quantify such an interpretation, we first compute the heat map of the original input image, which represents the information distribution \cite{zhou2016learning}.
We use the output feature maps of the last convolutional layer in the network, and add them together on the channel dimension.
Then we scale the sumed feature map to the resolution of the input image by bilinear interpolation.
Each pixel value $H_{i,j}$ in the heat map $H$ represents the importance of this pixel in the input image.
Finally, we compute the receptive field of a feature map on the heat map.
As shown in Fig. \ref{fig:exp}, the red part in a receptive field can quantify the interpretation of the corresponding feature map.
To this end, the information contained in a feature map can be viewed as the sum of the products between the elements of the mask and the heat map:

\vspace{-.5em}
\begin{small}
\begin{equation}
\begin{aligned}
\sum_{i=1}^{H_{im}}\sum_{j=1}^{W_{im}} H_{i,j} M_{i,j},
\end{aligned}
\label{exp:1}
\end{equation}
\end{small}

\vspace{-.5em}
\noindent where $H_{im}$ and $W_{im}$ denote the resolution (height and width) of the input image, and $M$ is the binary mask generated by the feature map. Eq. \ref{exp:1} can be rewritten as:

\vspace{-.7em}
\begin{small}
\begin{equation}
\begin{aligned}
 \mathbb{I}\{M=1\} \overline{\rm{H}},
\end{aligned}
\label{exp:2}
\end{equation}
\end{small}

\vspace{-1.5em}
\noindent where $\mathbb{I}\{M=1\}$ is the number of the elements in $M$ with value 1, which can be viewed as the area of the receptive field that depends on the sparsity of the feature map, and $\overline{\rm{H}}$ is the average of all entry $H_{i,j}$ whose corresponding element in $M$ is 1.
The heat map represents the information distribution in the input image.
The higher the value of an element in $H_{i,j}$ is, the more information this element contains.
Therefore, $\overline{\rm{H}}$ can represent the information richness in the feature map. Eq. \ref{exp:2} indicates that the importance of a feature map depends on its sparsity and information richness.
However, if we simply use Eq. \ref{exp:2} to compute the importance of each feature map, it suffers from heavy computation cost, since we need to compute all the feature maps with respect to the entire training dataset to obtain a comparative generalized result.

%------------------------------------------------------------------------
\section{Proposed Method}
\label{sec:4}

To handle the above issue, we introduce the kernel sparsity and entropy (KSE), which serves as an indicator to represent the sparsity and information richness of input feature maps in a feature-agnostic manner.
It is generic, and can be used to compress fully-connected layers by treating them as $1\times 1$ convolutional layers.

Generally, a convolutional layer transforms an input tensor $\mathcal{X} \in \mathbb{R}^{C\times H_{in}\times W_{in}}$ into an output tensor $\mathcal{Y} \in \mathbb{R}^{N\times H_{out}\times W_{out}}$ by using the filters $\mathcal{W} \in \mathbb{R}^{N \times C \times K_h \times K_w}$.
Here, $C$ is the number of the input feature maps, $N$ is the number of the filters, and $K_h$ and $K_w$ are the height and width of a filter, respectively.
The convolution operation can be formulated as follows:

\vspace{-.7em}
\begin{small}
\begin{equation}
  \label{method:eq.1}
\begin{aligned}
Y_n = \sum_{c=1}^{C} W_{n, c} * X_{c},
\end{aligned}
\end{equation}
\end{small}

\vspace{-.7em}
\noindent where $*$ represents the convolution operation, and $X_c$ and $Y_c$ are the channels (feature maps) of $\mathcal{X}$ and $\mathcal{Y}$, respectively. For simplicity, the biases are omitted for easy presentation. For an input feature map $X_c$, we call the set $\{ W_{n,c} \}_{n=1}^{N}$ the corresponding $2$D kernels of $X_c$.

%------------------------------------------------------------------------
\subsection{Kernel Sparsity}
\label{sec:4.1}
We measure the sparsity of an input feature map (\emph{i.e.}\ $\mathbb{I}\{M=1\}$) by calculating the sparsity of its corresponding $2$D kernels, \emph{i.e.,}\ the sum of their $\ell_1$-norms $ \sum_{n} | W_{n, c} |$.
Although these kernels do not participate in generating the input feature map, the sparsity between input feature map and its corresponding $2$D kernels is closely related.

During training, the update of the $2$D kernel $W_{n,c}$ depends on the gradient $\frac{\partial L}{\partial W_{n,c}}$ and the weight decay $R(W_{n,c})$:

\vspace{-.5em}
\begin{small}
\begin{equation}
\begin{aligned}
W_{n,c}^{(t+1)} &= W_{n,c}^{(t)} - \eta \frac{\partial L}{\partial W_{n,c}^{(t)}} - \frac{\partial R(W_{n,c}^{(t)})}{\partial  W_{n,c}^{(t)}} \\
&= W_{n,c}^{(t)} - \eta \frac{\partial L}{\partial Y_{n}^{(t)}} X_{c}^{(t)} - \frac{\partial R(W_{n,c}^{(t)})}{\partial  W_{n,c}^{(t)}} ,
\end{aligned}
\end{equation}
\end{small}

\vspace{-.5em}
\noindent where $L$ represents the loss function and $\eta$ is the learning rate.
If the input feature map $X_c^{(t)}$ is sparse, the kernel's gradient is relatively small, and the update formula becomes:

\vspace{-.5em}
\begin{small}
\begin{equation}
\begin{aligned}
W_{n,c}^{(t+1)} \approx W_{n,c}^{(t)} - \frac{\partial R(W_{n,c}^{(t)})}{\partial  W_{n,c}^{(t)}}.
\end{aligned}
\end{equation}
\end{small}

\vspace{-.5em}
\noindent Note that $|W_{n,c}^{(t+1)}|\rightarrow 0$ with the iterations if $R(W_{n,c}^{(t)})$ is defined based on the $\ell_2 $-regularization, which may make the kernel being sparse \cite{zhang2015sparse}.
Thus, the kernel corresponding to the sparse input feature map may be sparse during training.
Therefore, for the $c$-th input feature map, we define its sparsity as:

\vspace{-.5em}
\begin{small}
\begin{equation}
\begin{aligned}
s_c = \sum_{n=1}^{N} |W_{n, c}|.
\end{aligned}
\end{equation}
\end{small}

\vspace{-.5em}
To the best of our knowledge, we are the first to build the relationship between the input feature maps (rather than the output feature maps) and the kernels in terms of sparsity.
Note that this sparsity relation has been verified in our experiments shown in Section \ref{sec:sparsity}.

%------------------------------------------------------------------------
\subsection{Kernel Entropy}
\label{sec:4.2}
For a convolutional layer, if the convolutional results from a single input feature map are more diversified, this feature map contains more information and should be maintained in the compressed network.
%
%When the $2$D kernels corresponding an input feature map become more dense, their convolution results will be similar, which indicates that this input feature map does not provide much useful information.
In this case, the distribution of the corresponding 2D kernels is more complicated.
Therefore, we propose a new concept called kernel entropy to measure the information richness of an input feature map.

We first construct a nearest neighbor distance matrix $A_c$ for the $2$D kernels corresponding to the $c$-th input feature map. For each row $i$ and col $j$, if $W_{i,c}$ and $W_{j,c}$ are ``close''\footnote{Let $W_{i,c}^{'}$ and $W_{j,c}^{'}$ be two vectors formed by the elements of $W_{i,c}$ and $W_{j,c}$, respectively. Then the distance between $W_{i,c}$ and $W_{j,c}$ is defined as the distance between $W_{i,c}^{'}$ and $W_{j,c}^{'}$.}, \emph{i.e.,}\ $W_{j,c}$ is among the $k$ nearest neighbours of $W_{i,c}$, then $A_{c_{i,j}}= \| W_{i,c} - W_{j,c} \|$, and $A_{c_{i,j}}=0$ otherwise.
We set $k$ to $5$ empirically, which can achieve good results.
Then we calculate the density metric of each kernel by the number of instances located in the neighborhood of this kernel:

\vspace{-.7em}
\begin{small}
\begin{equation}
\begin{aligned}
dm(W_{i,c}) = \sum_{j=1}^{N} A_{c_{i,j}}.
\end{aligned}
\end{equation}
\end{small}

\vspace{-.5em}
\noindent The larger the density metric of $W_{i,c}$ is, the smaller the density of $W_{i,c}$ is, \emph{i.e.,}\ the kernel is away from the others, and the convolutional result using $W_{i,c}$ becomes more different.
Hence, we define the kernel entropy to measure the complexity of the distribution of the 2D kernels:

\vspace{-.5em}
\begin{small}
\begin{equation}
\begin{aligned}
e_c = -\sum_{i=1}^{N} \frac{dm(W_{i,c})}{d_c} log_2 \frac{dm(W_{i,c})}{d_c},
\end{aligned}
\end{equation}
\end{small}

\noindent where $d_c = \sum_{i=1}^{N} dm(W_{i,c})$.
The smaller the kernel entropy is, the more complicated the distribution of the $2$D kernels is, and the more diverse the kernels are. In this case, the features extracted by these kernels have greater difference.
Therefore, the corresponding input feature map provides more information to the network.

%------------------------------------------------------------------------
\subsection{Definition of the KSE Indicator}
\label{sec:4.3}
As discussed in Section \ref{sec:3}, the feature map importance depends on two parts, the sparsity and the information richness.
Upon this discovery, we first use the min-max normalization $s_c$ and $e_c$ into $[0, 1]$ to make them in the same scale.
Then we combine the kernel sparsity and entropy to measure the overall importance of an input feature map by:

\vspace{-.5em}
\begin{equation}
\begin{aligned}
v_c = \sqrt{\frac{s_c}{1 + \alpha e_c}},
\end{aligned}
\end{equation}

\noindent where $\alpha$ is a parameter to control the balance between the sparsity and entropy, which is set to 1 in this work.
We call $v_c$ the KSE indicator that measures the interpretability and importance of the input feature map.
We further use the min-max normalization to rescale the indicators to $[0, 1]$ based on all the input feature maps in one convolutional layer.

%------------------------------------------------------------------------
\subsection{Kernel Clustering}
\label{sec:4.4}

To compress kernels, previous channel pruning methods divide channels into two categories based on their importance, \emph{i.e.,}\ important or unimportant ones.
Thus, for an input feature map, its corresponding $2$D kernels are either all kept or all deleted, which is a coarse compression.
In our work, through clustering, we develop a fine-grained compression scheme to reduce the number of the kernels where after pruning, the number is an integer between $0$ and $N$ (but not just $0$ or $N$ as in the previous methods).

First, we decide the number of kernels required for their corresponding $c$-th input feature map as:

\begin{equation}
q_c = \left\{
\begin{array}{lcl}
0,                                        &  & \lfloor v_cG \rfloor=0, \\
N,                                        &  & \lceil v_cG \rceil = G, \\
\big\lceil\frac{N}{2^{G - \lceil v_cG \rceil + T}}\big\rceil, &  & \text{otherwise},

\end{array}
\right.
\end{equation}

\noindent where $G$ controls the level of compression granularity.
A larger $G$ results in a finer granularity.
$N$ is the number of the original $2$D kernels and $T$ is a hyper-parameter to control the compression and acceleration ratios.

Second, to guarantee each output feature map contains the information from most input feature maps, we choose to cluster, rather than pruning, the $2$D kernels to reduce the kernel number.
It is achieved simply by the k-means algorithm with the number of cluster centroids equal to $q_c$.
Thus, the $c$-th input feature map generates $q_c$ centroids (new $2$D kernels)
$\{B_{i,c} \in \mathbb{R}^{K_h\times K_w}\}_{i=1}^{q_c}$
and an index set
$\big\{I_{n,c} \in \{ 1, 2,...,q_c \}\big\}_{n=1}^{N}$
to replace the original $2$D kernels
$\{W_{n,c} \in \mathbb{R}^{K_h\times K_w}\}_{n=1}^{N}$.
For example, $I_{1,c}=2$ denotes that the first original kernel is classified to the second cluster $B_{2,c}$.
When $q_c=0$, the $c$-th input feature map is considered as unimportant, it and all its corresponding kernels are pruned.
In the other extreme case where $q_c=N$, this feature map is considered as most important for the convolutional layer, and all its corresponding kernels are kept.

There are three steps in our training procedure.
(i) Pre-train a network on a dataset.
(ii) Compress the network (in all the convolutional and fully-connected layers) using the method proposed above and obtain $q_c$, $B_{i,c}$ and $I_{n,c}$.
(iii) Fine-tune the compressed network for a small number of epochs.
During the fine-tuning, we only update the cluster centroids.

In inference, our method accelerates the network by sharing the $2$D activation maps extracted from the input feature maps to reduce the computation complexity of convolution operations.
Note that here the activation maps (yellow planes in Fig. \ref{fig:short}) are not the output feature maps (orange planes in Fig. \ref{fig:short}).
As shown in Fig. \ref{fig:short}, we split the convolutional operation into two parts, $2$D convolution and channel fusion.
In $2$D convolution, the responses from each input feature map are computed simultaneously to generate the $2$D activation maps.
The $c$-th input feature map corresponds to the $q_c$ $2$D kernels, which generates $q_c$ $2$D activation maps $Z_{i,c} = B_{i,c}*X_{c}$.
Then, in the channel addition, we compute $Y_n$ by summing their corresponding $2$D activation maps:

\vspace{-1em}
\begin{small}
\begin{equation}
\begin{aligned}
Y_n = \sum_{c=1}^{C} Z_{I_{n,c}, c}.
\end{aligned}
\end{equation}
\vspace{-1.5em}
\end{small}
%------------------------------------------------------------------------
\section{Experiments}
\label{sec:5}

We have implemented our method using Pytorch \cite{paszke2017automatic}.
The effectiveness validation is performed on three datasets, CIFAR-10, Street View House Numbers (SVHN), and ImageNet ILSVRC 2012.
CIFAR-10 has $32 \times 32$ images from 10 classes.
The training set contains 50,000 images and the test set contains 10,000 images.
The SVHN dataset has 73,257 $32\times 32$ color digit images in the training set and 26,032 images in the test set.
ImageNet ILSVRC 2012 consists of 1.28 million images for training and 50,000 images for validation over 1,000 classes.

All networks are trained using stochastic gradient descent (SGD) with momentum set to 0.9.
On CIFAR-10 and SVHN, we respectively train the networks for 200 and 20 epochs using the mini-batch size of 128.
The initial learning rate is 0.01 and is multiplied by 0.1 at 50\% of the total number of epochs.
On ImageNet, we train the networks for 21 epochs with the mini-batch size of 64, and the initial learning rate is 0.001 which is divided by 10 at epoch 7 and 14.
Because the first layer has only three input channels and the last layer is a classifier in ResNet and DenseNet, we do not compress the first and the last layers of the networks.

In our method, we use $T$ to control the compression and acceleration ratios, which determines the number of 2D kernels after compression.
In experiments, we set $T$ to $0$ for ResNet-56, DenseNet-40-12 and DenseNet-100-12 on both CIFAR-10 and SVHN to achieve better accuracies, and set $T$ to $1$ for ResNet-50 and all the DenseNets on ImageNet 2012 to achieve lower compression ratios.

\begin{figure}[t]
\begin{center}
\includegraphics[width=0.49\linewidth]{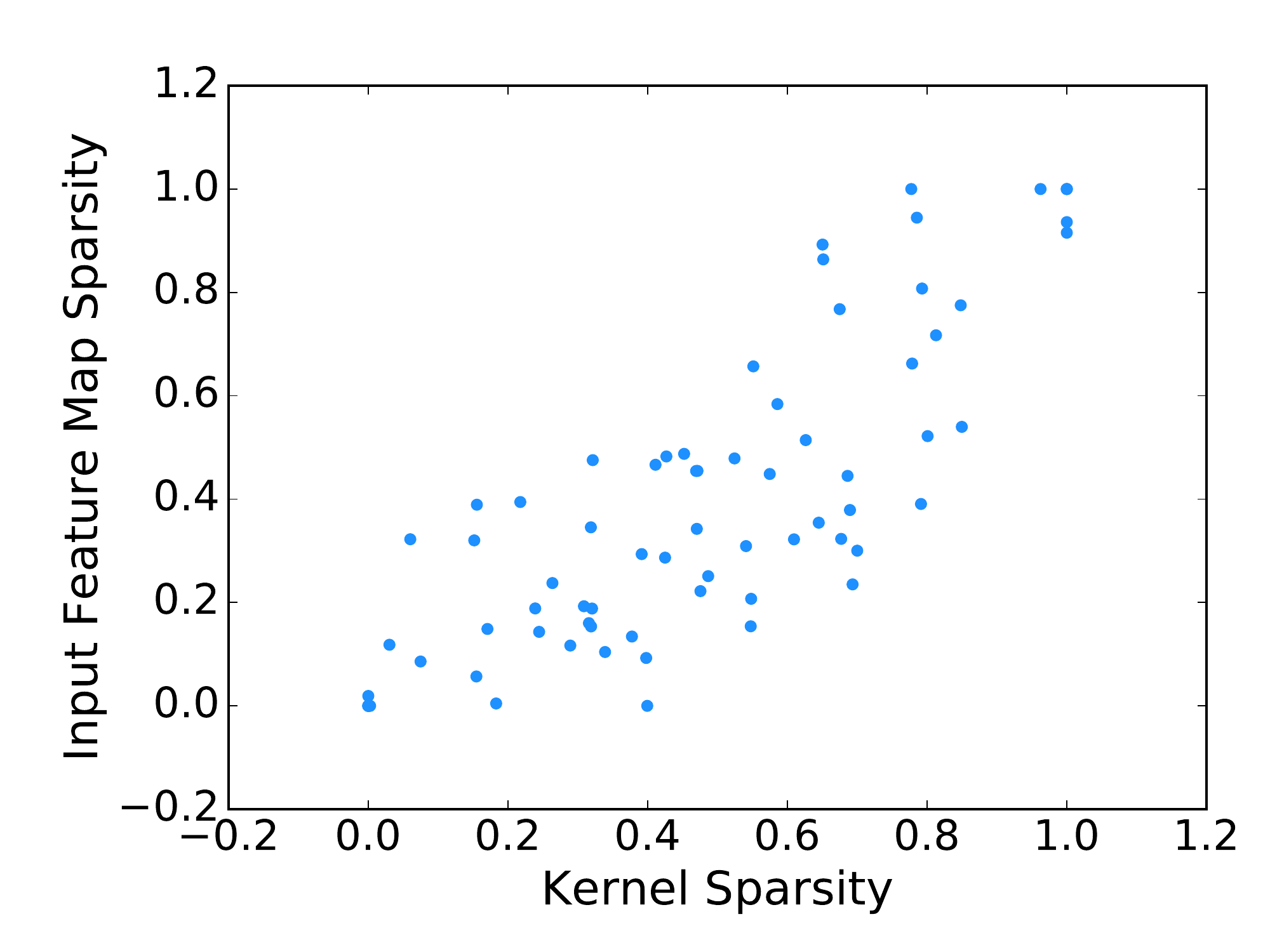}
\includegraphics[width=0.49\linewidth]{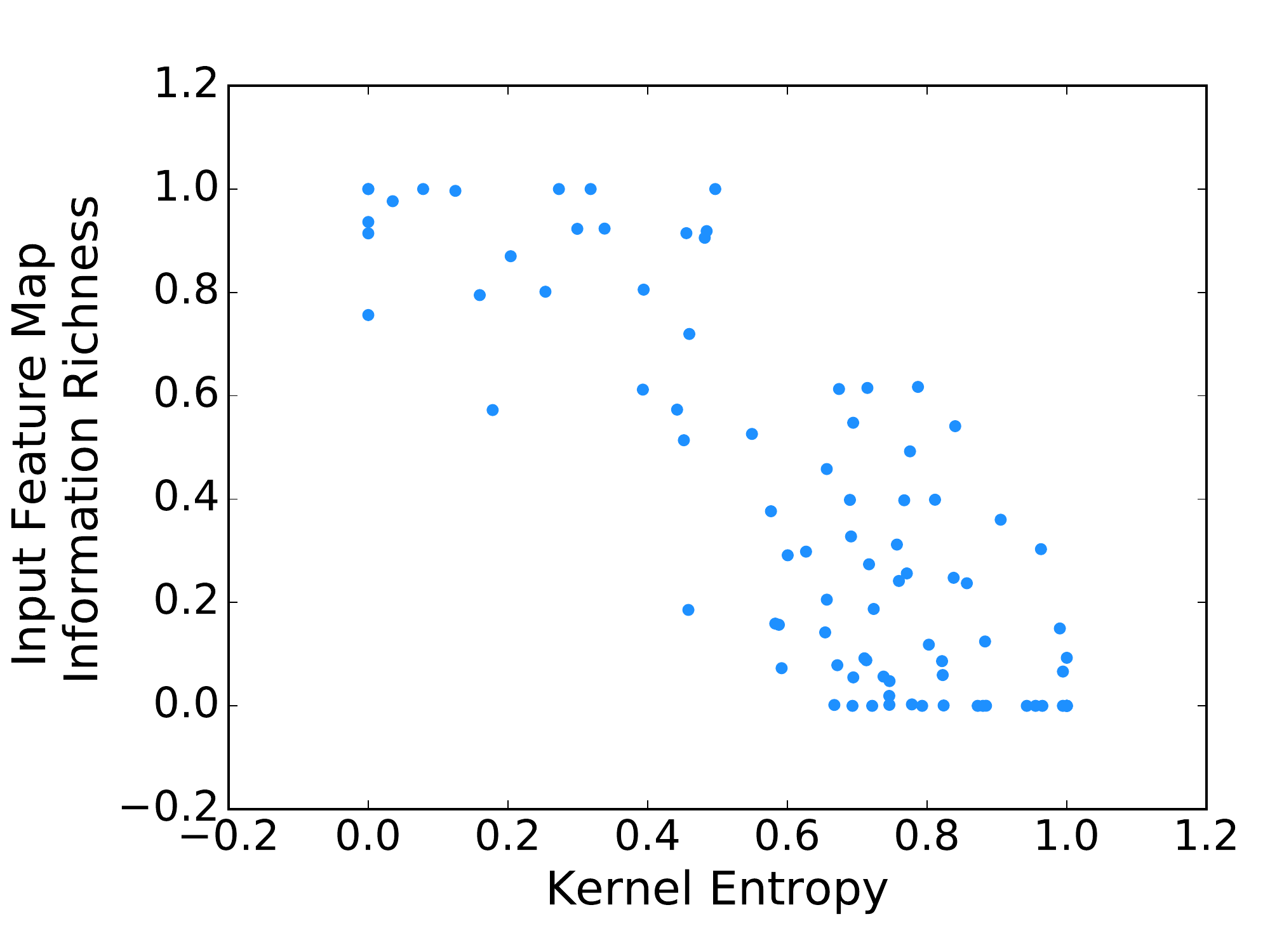}
\end{center}
\vspace{-.5em}
   \caption{Left: Relationship between the sparsity of the input feature maps and their corresponding kernel sparsity. Right: Relationship between the information richness of the input feature maps and their corresponding kernel entropy.}
   \vspace{-1em}
\label{fig:KSE}
\end{figure}

%------------------------------------------------------------------------
\subsection{Compression and Acceleration Ratios}

In this section, we analyze the compression and acceleration ratios.
For a convolutional layer, the size of the original parameters is $N\times C\times K_h\times K_w$ and each weight is assumed to require 32 bits.
We store $q_c$ cluster centroids, each weight of which again requires 32 bits.
Besides, each index per input feature map takes $log_2 q_c$ bits.
The size of each centroid is $K_h\times K_w$, and each input feature map needs $N$ indices for the correspondence.
Therefore, we can calculate the compression ratio $r_{comp}$ for each layer by:

\begin{small}
  \vspace{-.5em}
\begin{equation}
  \label{eq:compress}
\begin{aligned}
r_{comp} = \frac{NCK_hK_w}{\sum\limits_{c}(q_cK_hK_w+N\frac{log_2q_c}{32})}.
\end{aligned}
\vspace{-1em}
\end{equation}
\end{small}

We can also accelerate the compressed network based on the sharing of the convolution results.
As mentioned in Section \ref{sec:4.4}, we compute all the intermediate features from the input feature maps, and then add the corresponding features for each output feature map.
The computation is mainly consumed by convolution operations.
Thus, the theoretical acceleration ratio $r_{acce}$ of each convolutional layer is computed by:

\begin{small}
  \vspace{-.5em}
\begin{equation}
  \label{eq:accelerate}
\begin{aligned}
r_{acce} \simeq \frac{NCH_{out}W_{out}K_hK_w}{\sum\limits_{c}q_cH_{out}W_{out}K_hK_w} = \frac{NC}{\sum\limits_{c}q_c}.
\end{aligned}
\vspace{-1em}
\end{equation}
\end{small}

We can also calculate the compression and acceleration ratios on fully-connected layers by treating them as $1 \times 1$ convolutional layers.

%------------------------------------------------------------------------
\subsection{Relationship between Input Feature Maps and their Corresponding Kernels}
\label{sec:sparsity}

We calculate the sparsity $\mathbb{I}\{M=1\}$ and information richness $\overline{\rm{H}}$ of the input feature maps, and their corresponding kernel sparsity and entropy.
The relationships are shown in Fig. \ref{fig:KSE} where the feature maps are a random subset of all the feature maps from ResNet-56 on CIFAR-10.
We can see that the sparsity of the input feature maps and the kernel sparsity increase simultaneously, while the information richness of the input feature maps decreases as the kernel entropy increases.

We can further use the Spearman correlation coefficient $\rho$ to quantify these relationships:

\begin{small}
  \vspace{-.5em}
\begin{equation}
  \label{eq:spearman}
\begin{aligned}
 \rho = \frac{\sum_i(x_i - \overline{x})(y_i - \overline{y})}{\sqrt{\sum_i(x_i - \overline{x})^2 \sum_i (y_i - \overline{y})^2}},
\end{aligned}
  \vspace{-.5em}
\end{equation}
\end{small}

\noindent where $\overline{x}$ and $\overline{y} $ are the averages of the random variables $x$ and $y$, respectively.
The correlation coefficient for the relationship on the left of Fig. \ref{fig:KSE} is $0.833$, while the correlation coefficient for the relationship on the right of Fig. \ref{fig:KSE} is $-0.826$. These values confirm the positive correlation for the first and the negative correlation for the second.

%------------------------------------------------------------------------
\subsection{Comparison with State-of-the-Art Methods}
\label{experiment:compare}

\textbf{CIFAR-10}. We compare our method with \cite{he2017channel,li2016pruning} on ResNet-56, and with \cite{huang2018condensenet, liu2017learning, son2018clustering} on DenseNet.
For ResNet-56, we set $G$ to two values ($4$ and $5$) to compress the network.
For DenseNet-40-12 and DenseNet-BC-100-12, we set $G$ to another two values ($3$ and $6$).
As shown in Table \ref{tab:1} and Table \ref{tab:2}, our method achieves the best results, compared to the filter pruning methods \cite{he2017channel,li2016pruning} on ResNet-56 and the channel pruning methods \cite{huang2018condensenet,liu2017learning} on DenseNet.
Moreover, our method also achieves better results than other kernel clustering methods \cite{son2018clustering} on DenseNet-BC-100-12.
Compared to Son \emph{et al.} \cite{son2018clustering}, our KSE method can not only compress the $3\times 3$ convolutional layers, but also the $1 \times 1$ convolutional layers to obtain a more compact network.

\begin{table}
  \footnotesize
\begin{center}
\begin{tabular}{|p{3.1cm}<{\centering}|p{1.4cm}<{\centering}|p{1.4cm}<{\centering}|p{0.8cm}<{\centering}|}
\hline
\multirow{2}*{Model} & FLOPs   ($r_{acce}$) & \#Param.    ($r_{comp}$) & Top-1 Acc\% \\
\hline
ResNet-56$_{baseline}$ & 125M(1.0$\times$) & 0.85M(1.0$\times$) & 93.03\\
\hline
ResNet-56-pruned-A \cite{li2016pruning} & 112M(1.1$\times$) & 0.77M(1.1$\times$) & 93.10\\
ResNet-56-pruned-B \cite{li2016pruning} & 90M(1.4$\times$) & 0.73M(1.2$\times$) & 93.06\\
\hline
ResNet-56-pruned \cite{he2017channel} & 62M(2.0$\times$) & - & 91.80\\
\hline
KSE (G=4) & 60M(2.1$\times$) & 0.43M(2.0$\times$) & \textbf{93.23}\\
KSE (G=5) & 50M((2.5$\times$) & 0.36M(2.4$\times$) & \textbf{92.88}\\
\hline
\end{tabular}
\end{center}
\caption{Results of ResNet-56 on CIFAR-10. In all tables and figures, M/B means million/billion.}
\label{tab:1}
\end{table}

\begin{table}
  \footnotesize
\begin{center}
\begin{tabular}{|p{3.1cm}<{\centering}|p{1.4cm}<{\centering}|p{1.4cm}<{\centering}|p{0.8cm}<{\centering}|}
\hline
\multirow{2}*{Model} & FLOPs   ($r_{acce}$) & \#Param.    ($r_{comp}$) & Top-1 Acc\% \\
\hline
DenseNet-40$_{baseline}$ & 283M(1.0$\times$) & 1.04M(1.0$\times$) & 94.81\\
\hline
DenseNet-40 (40\%) \cite{liu2017learning} & 190M(1.5$\times$) & 0.66M(1.6$\times$) & 94.81\\
DenseNet-40 (70\%) \cite{liu2017learning} & 120M(2.4$\times$) & 0.35M(3.0$\times$) & 94.35\\
\hline
KSE (G=3) & 170M(1.7$\times$) & 0.63M(1.7$\times$) & \textbf{94.81}\\
KSE (G=6) & 115M(2.5$\times$) & 0.39M(2.7$\times$) & \textbf{94.70}\\
\hline
\multicolumn{1}{|l}{} & \multicolumn{1}{l}{} & \multicolumn{1}{l}{} & \multicolumn{1}{l|}{} \\[-2 ex]
\hline
DenseNet-BC-100$_{baseline}$ & 288M(1.0$\times$) & 0.75M(1.0$\times$) & 95.45\\
\hline
DenseNet-PC128N \cite{son2018clustering} & 212M(1.4$\times$) & 0.50M(1.5$\times$) & 95.43\\
\hline
CondenseNet$^{light}$-94 \cite{huang2018condensenet} & 122M(2.4$\times$) & 0.33M(2.3$\times$) & 95.00\\
\hline
KSE (G=3) & 159M(1.8$\times$) & 0.45M(1.7$\times$) & \textbf{95.49}\\
\hline
KSE (G=6) & 103M(2.8$\times$) & 0.31M(2.4$\times$) & \textbf{95.08}\\
\hline
\end{tabular}
\end{center}
\caption{Results of DenseNet on CIFAR-10.}
\label{tab:2}
\end{table}

\begin{table}
  \footnotesize
\begin{center}
\begin{tabular}{|p{3.1cm}<{\centering}|p{1.4cm}<{\centering}|p{1.4cm}<{\centering}|p{0.8cm}<{\centering}|}
\hline
\multirow{2}*{Model} & FLOPs   ($r_{acce}$) & \#Param.    ($r_{comp}$) & Top-1 Acc\% \\
\hline
DenseNet-40$_{baseline}$ & 283M(1.0$\times$) & 1.04M(1.0$\times$) & 98.17\\
\hline
DenseNet-40 (40\%) \cite{liu2017learning} & 185M(1.5$\times$) & 0.65M(1.6$\times$) & 98.21\\
DenseNet-40 (60\%) \cite{liu2017learning} & 134M(2.1$\times$) & 0.44M(2.4$\times$) & 98.19\\
\hline
KSE (G=4) & 147M(1.9$\times$) & 0.49M(2.1$\times$) & \textbf{98.27}\\
KSE (G=5) & 130M(2.2$\times$) & 0.42M(2.5$\times$) & \textbf{98.25}\\
\hline
\end{tabular}
\end{center}
\caption{Results of DenseNet-40-12 on SVHN.}
\vspace{-1em}
\label{tab:3}
\end{table}

\textbf{SVHN}. We also evaluate the performance of KSE on DenseNet-40-12 on SVHN.
We set $G$ to two values, $4$ and $5$.
As shown in Table \ref{tab:3}, our method achieves better performance than the channel pruning method \cite{liu2017learning}.
For example, compared to He \emph{et al.} \cite{liu2017learning}, we obtain 0.06\% increase in Top-1 accuracy (98.25\% \emph{vs.} 98.19\%) with higher compression and acceleration ratios ($2.5\times$ and $2.2\times$ \emph{vs.} $2.4\times$ and $2.1\times$).

\textbf{ImageNet 2012}. We set $G$ to $4$ and $5$, and compare our KSE method with three state-of-the-art methods \cite{he2017channel, lin2018accelerating,luo2017thinet}.
As show in Table \ref{tab:4}, Our method achieves the best performance with only a decrease of 0.35\% in Top-5 accuracy by a factor of $2.9 \times$ compression and $4.7 \times$ speedup.
These state-of-the-art methods perform worse mainly because they use dichotomy to compress networks, \emph{i.e.,} prune or keep the filters/channels, which leads to the loss of some important information in the network.
Besides, filter pruning methods like \cite{lin2018accelerating,luo2017thinet} cannot be applied to some convolutional layers due to the dimensional mismatch problem on ResNets with a multi-branch architecture.

\begin{table}
  \footnotesize
\begin{center}
\begin{tabular}{|p{2.3cm}<{\centering}|p{1.35cm}<{\centering}|p{1.5cm}<{\centering}|p{0.68cm}<{\centering}|p{0.68cm}<{\centering}|}
\hline
\multirow{2}*{Model}  & FLOPs   ($r_{acce}$) & \#Param.    ($r_{comp}$) & Top-1 Acc\% & Top-5 Acc\% \\
\hline
ResNet-50$_{baseline}$ & 4.10B(1.0$\times$) & 25.56M(1.0$\times$) & 76.15 & 92.87 \\
\hline
GDP-0.6 \cite{lin2018accelerating} & 1.88B(2.2$\times$) & - & 71.89 & 90.71\\
\hline
ResNet-50(2$\times$) \cite{he2017channel} & 2.73B(1.5$\times$) & - & 72.30 & 90.80\\
\hline
ThiNet-50 \cite{luo2017thinet} & 1.71B(2.4$\times$) & 12.38M(2.0$\times$) & 71.01 & 90.02\\
ThiNet-30 \cite{luo2017thinet} & 1.10B(3.7$\times$) & 8.66M(3.0$\times$) & 68.42 & 88.30\\
\hline
KSE (G=5) & 1.09B(3.8$\times$) & 10.00M(2.6$\times$) & \textbf{75.51} & \textbf{92.69}\\
KSE (G=6) & 0.88B(4.7$\times$) & 8.73M(2.9$\times$) & \textbf{75.31} & \textbf{92.52}\\
\hline
\end{tabular}
\end{center}
\caption{Results of ResNet-50 on ImageNet2012.}
\label{tab:4}
\end{table}

\begin{figure}[t]
\begin{center}
  \vspace{-1em}
\includegraphics[width=0.49\linewidth]{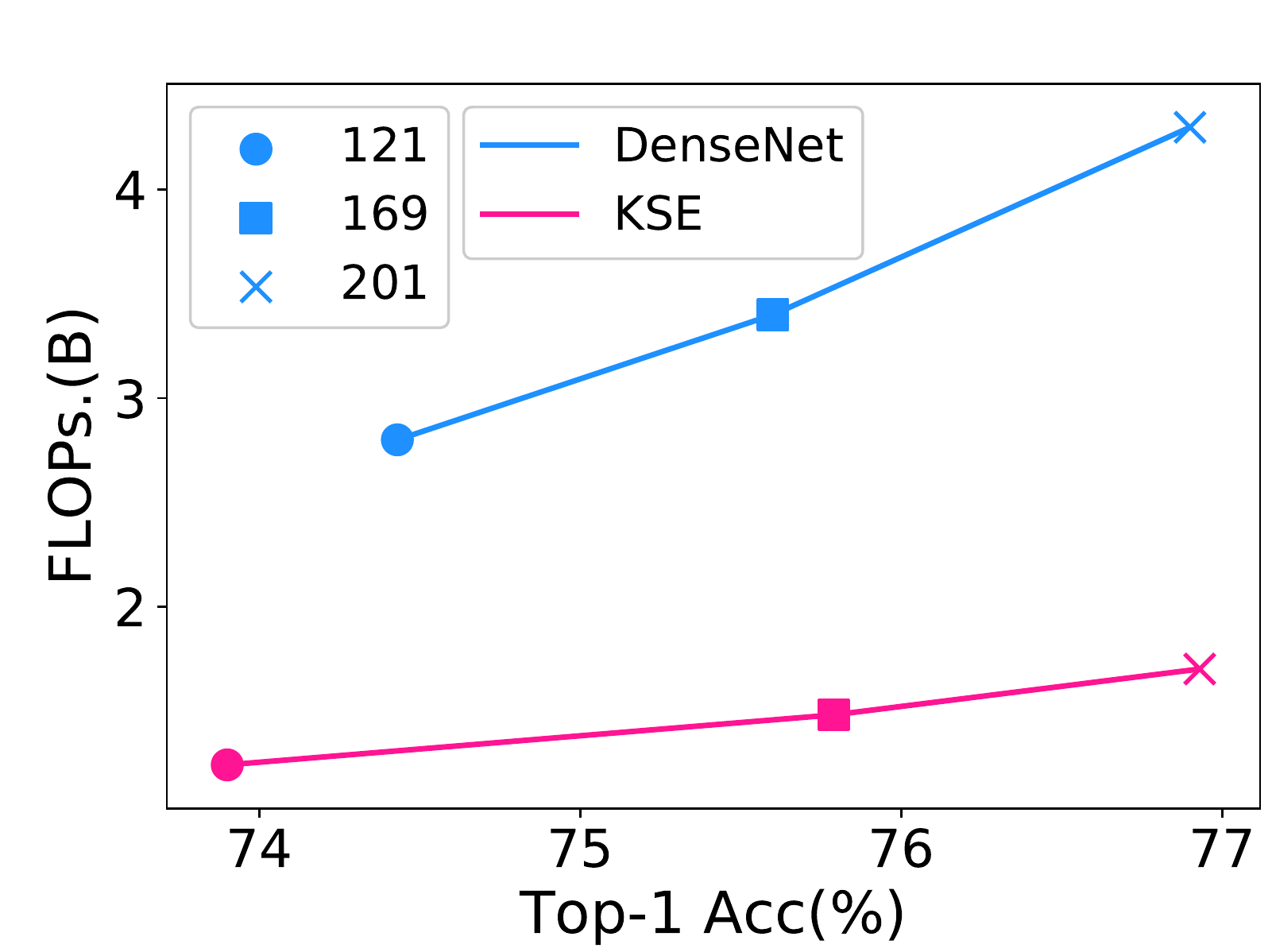}
\includegraphics[width=0.49\linewidth]{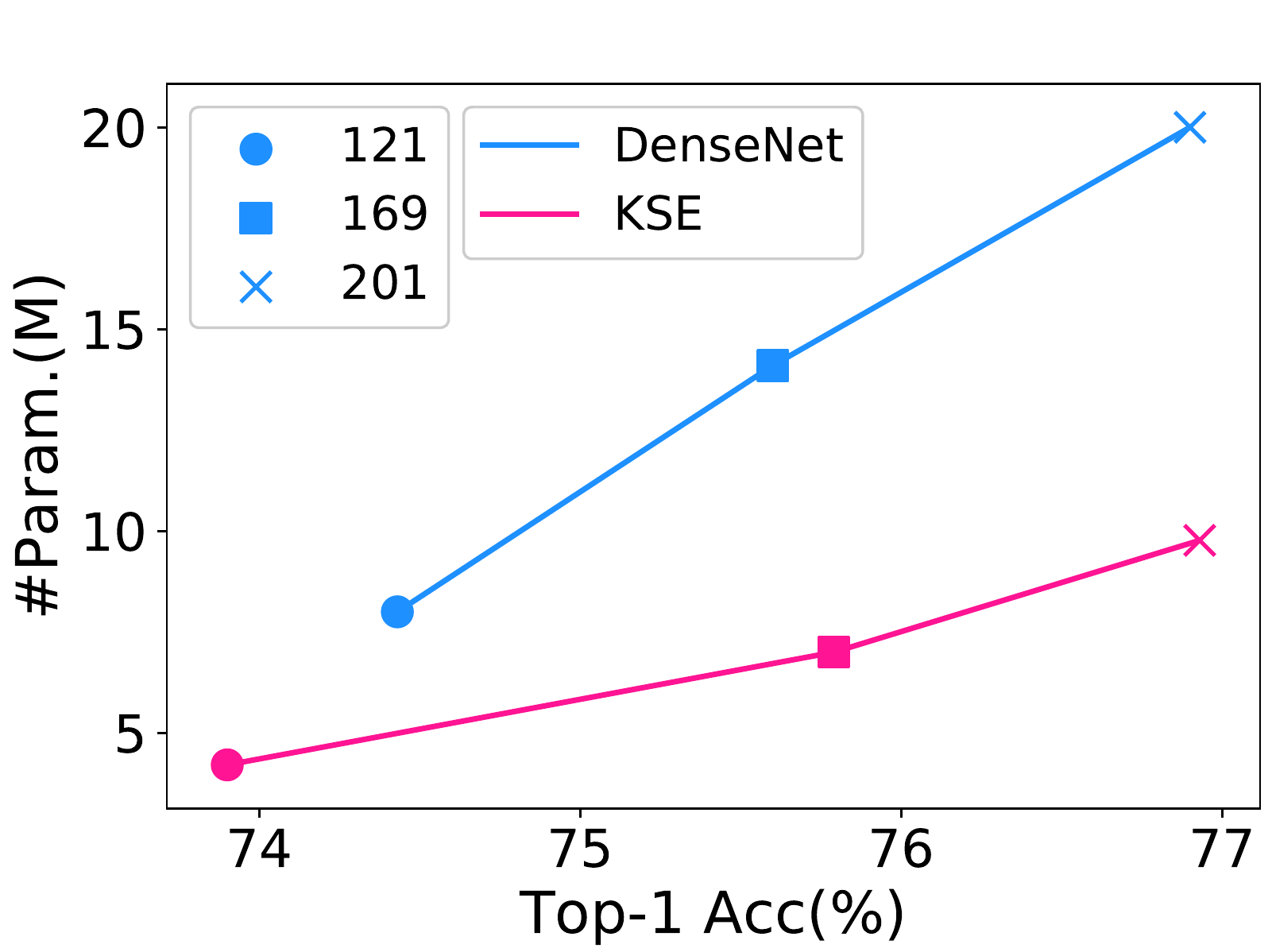}
\end{center}
\vspace{-.5em}
   \caption{Parameter and FLOPs amount comparison between the original DenseNets and the compressed DenseNets by our KSE.}
   \vspace{-1em}
\label{fig:densenet}
\end{figure}

For DenseNets, we set $G$ to 4, and compress them with three different numbers of layers, DenseNet-121, DenseNet-169, and DenseNet-201.
As shown in Fig. \ref{fig:densenet}, the compressed networks by KSE is on par with the original networks, but achieves almost $2\times$ parameter compression and speedup.

\begin{figure}[t]
\begin{center}
%\fbox{\rule{0pt}{2in} \rule{0.9\linewidth}{0pt}}
\vspace{-1em}
\hspace{-2em}
\includegraphics[width=0.75\linewidth]{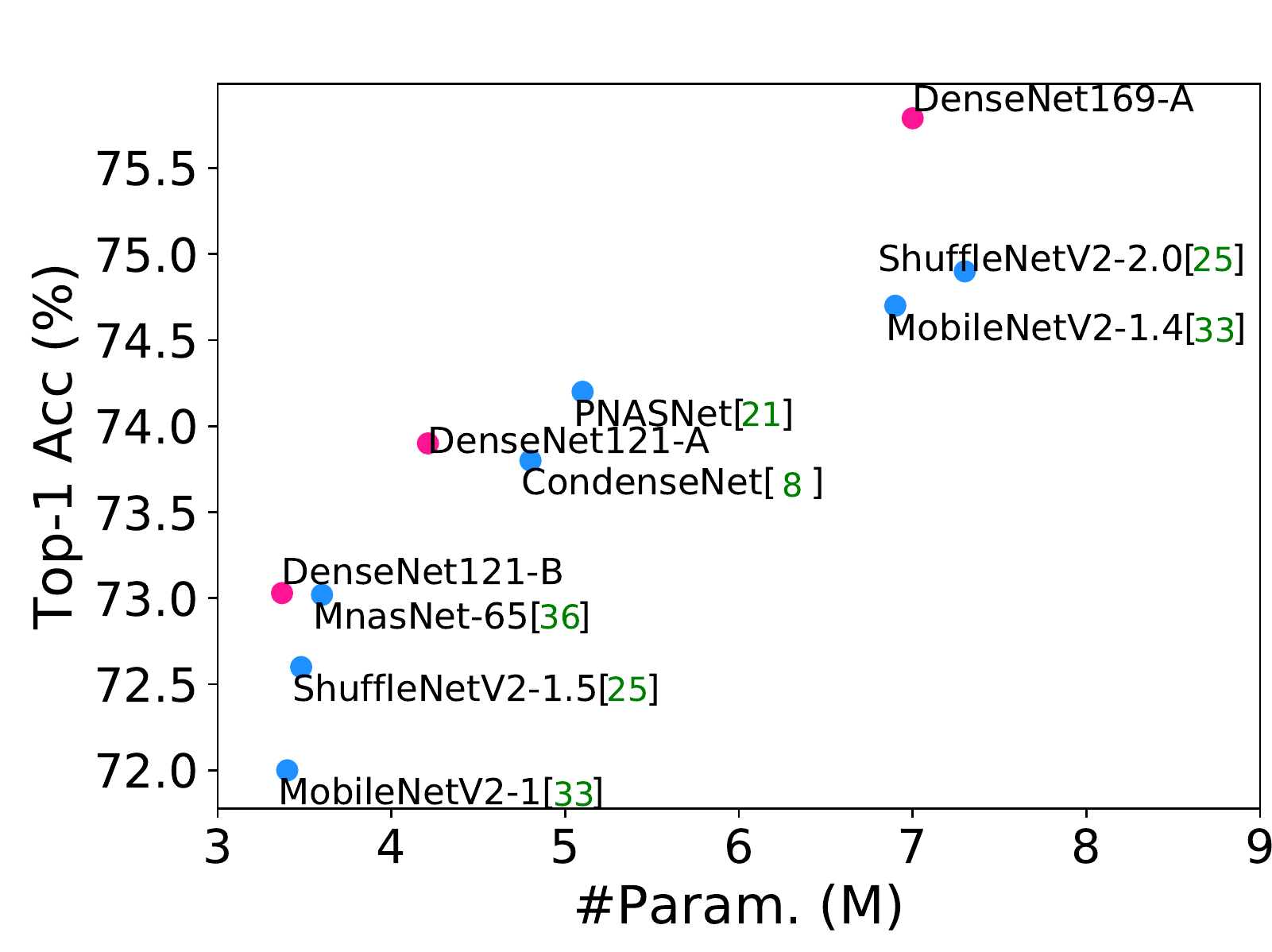}
\end{center}
\vspace{-.5em}
   \caption{Comparison of the compressed DenseNets (red circles) by our KSE and other compact networks (blue circles).}
\label{fig:densenet_compare}
\end{figure}

Recently, many compact networks \cite{ma2018shufflenet,sandler2018mobilenetv2} have been proposed to be used on mobile and embedded devices.
In addition, auto-search algorithms \cite{liu2017progressive, tan2018mnasnet} have been proposed to search the best network architecture by reinforcement learning.
We compare the compressed DenseNet-121 and DenseNet-169 by KSE with these methods \cite{huang2018condensenet, liu2017progressive,ma2018shufflenet, sandler2018mobilenetv2, tan2018mnasnet}.
In Fig. \ref{fig:densenet_compare}. `A' represents $G=4$ and `B' represents $G=3$.
We use KSE to compress DenseNets, which achieves more compact results.
For example, we obtain 73.03\% Top-1 accuracy with only 3.37M parameters on ImageNet 2012.
Our KSE uses different numbers of 2D kernels for different input feature maps to do convolution, which reduces more redundant kernels, compared to the complicated auto-search algorithms which only use the traditional convolutional layer.
Besides, the widely-used depth-wise convolution on MobileNet or ShuffleNet may cause significant information loss, due to only one 2D kernel is used to extract features from each feature map.

%------------------------------------------------------------------------
\subsection{Ablation Study}

The effective use of KSE, is related to $G$.
We select ResNet-56 and DenseNet-40 on CIFAR-10, and ResNet-50 and DenseNet-121 on ImageNet2012 to evaluate $G$.
Moreover, we analyze three different indicators.

%------------------------------------------------------------------------

\begin{figure}[t]
\begin{center}
%\fbox{\rule{0pt}{2in} \rule{0.9\linewidth}{0pt}}
\includegraphics[width=0.495\linewidth]{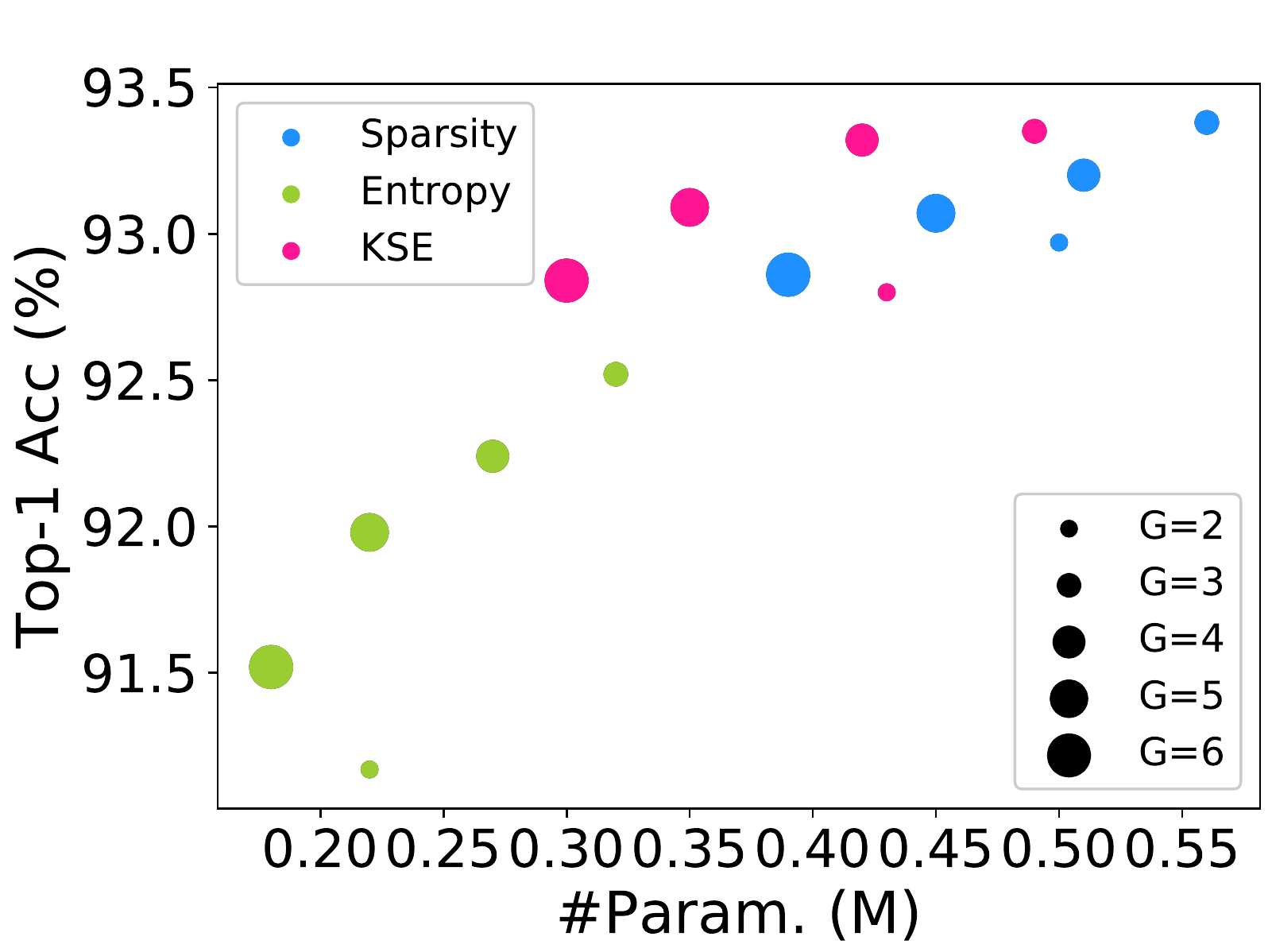}
\includegraphics[width=0.495\linewidth]{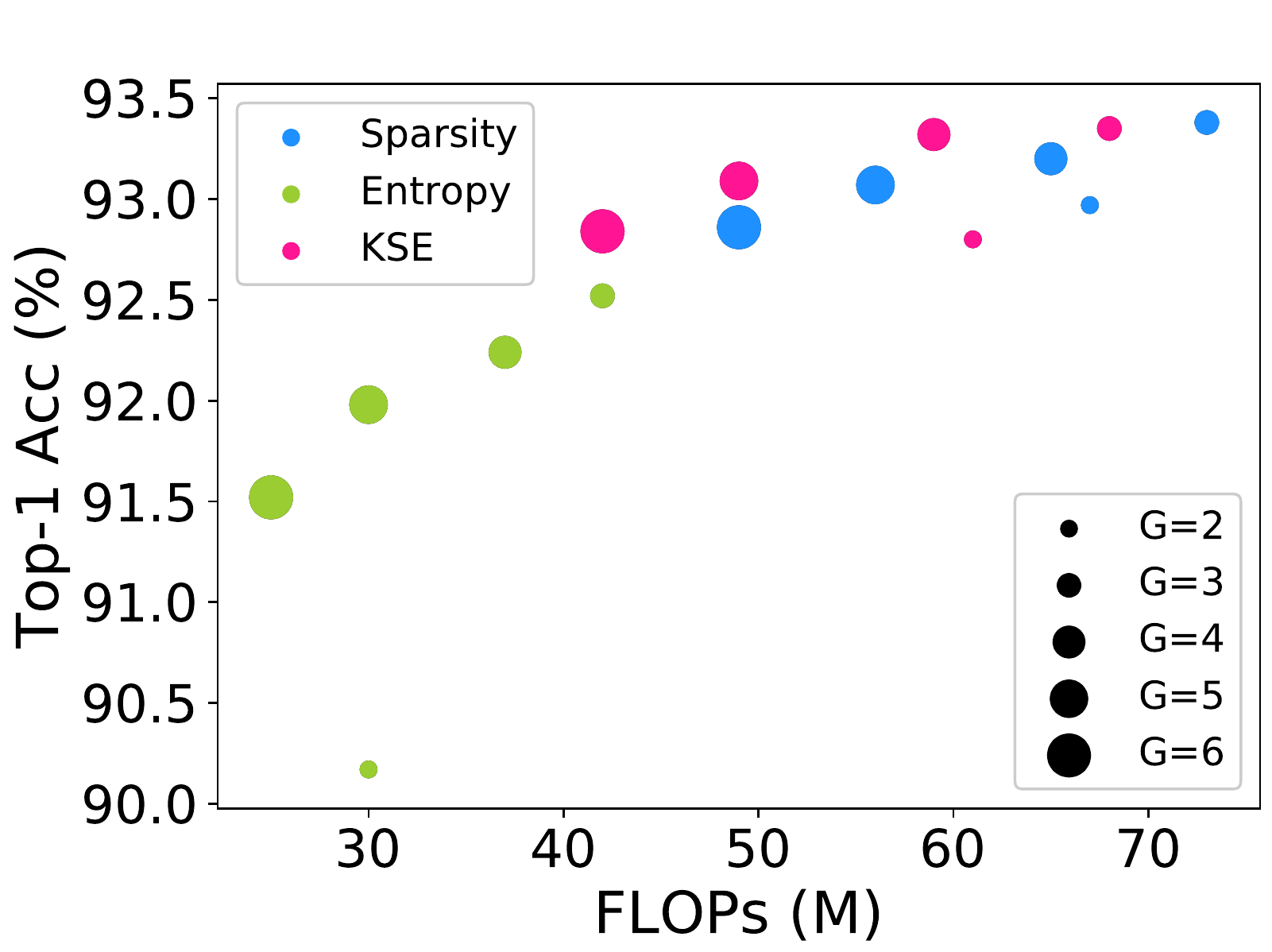}
\end{center}
\vspace{-.5em}
   \caption{Compression granularity and indicator analysis with ResNet-56 on CIFAR-10.}
   \vspace{-1em}
\label{fig:resnet_g}
\end{figure}

%------------------------------------------------------------------------
   \vspace{-1em}
\subsubsection{Effect of the Compression Granularity $G$}
   \vspace{-.5em}

In our kernel clustering, we use $G$ to control the level of compression granularity.
The results of different $G$ are shown in Fig. \ref{fig:resnet_g}.
Note that in this sub-section, only the red solid circles in Fig. \ref{fig:resnet_g} are concerned.
When $G=2$, the $2$D kernels corresponding to the $c$-th feature map are divided into two categories: $q_c=0$ or $q_c=N$ for pruning or keeping all the kernels.
It is a coarse-grained pruning method.
As $G$ increases, $q_c$ achieves various different values, which means to compress the $2$D kernels in a fine-grained manner.
In addition, the compression and acceleration ratios are also increased.
%
% To explain, the more detailed division of input feature map reduces the number of the whole cluster centroids, which results in large improvements on compression and acceleration ratio.
%
Compared to the coarse-grained pruning, the fine-grained pruning achieves much better results.
For example, when $G=4$, it achieves the same compression ratio as $G=2$ with 0.52\% Top-1 accuracy increase.

%------------------------------------------------------------------------
   \vspace{-1em}
\subsubsection{Indicator Analysis}
   \vspace{-.5em}

In this paper, we have proposed three concepts, kernel sparsity, kernel entropy, and KSE indicator.
In fact, all of them can be used as an indicator to judge the importance of the feature maps.
We next evaluate these three indicators on ResNet-56, with the proposed kernel clustering for compression.
As shown in Fig. \ref{fig:resnet_g}, compared to the indicators of kernel sparsity and kernel entropy, the KSE indicator achieves the best results for different compression granularities.
This is due to the fact that the kernel sparsity only represents the area of the receptive field of a feature map, and the density entropy of the 2D kernels only expresses the position information of a receptive field, which alone are not as effective as KSE to evaluate the importance of a feature map.

%------------------------------------------------------------------------
\subsection{Visualization Analysis}

\begin{figure}[t]
\begin{center}
  \includegraphics[width=1.0\linewidth]{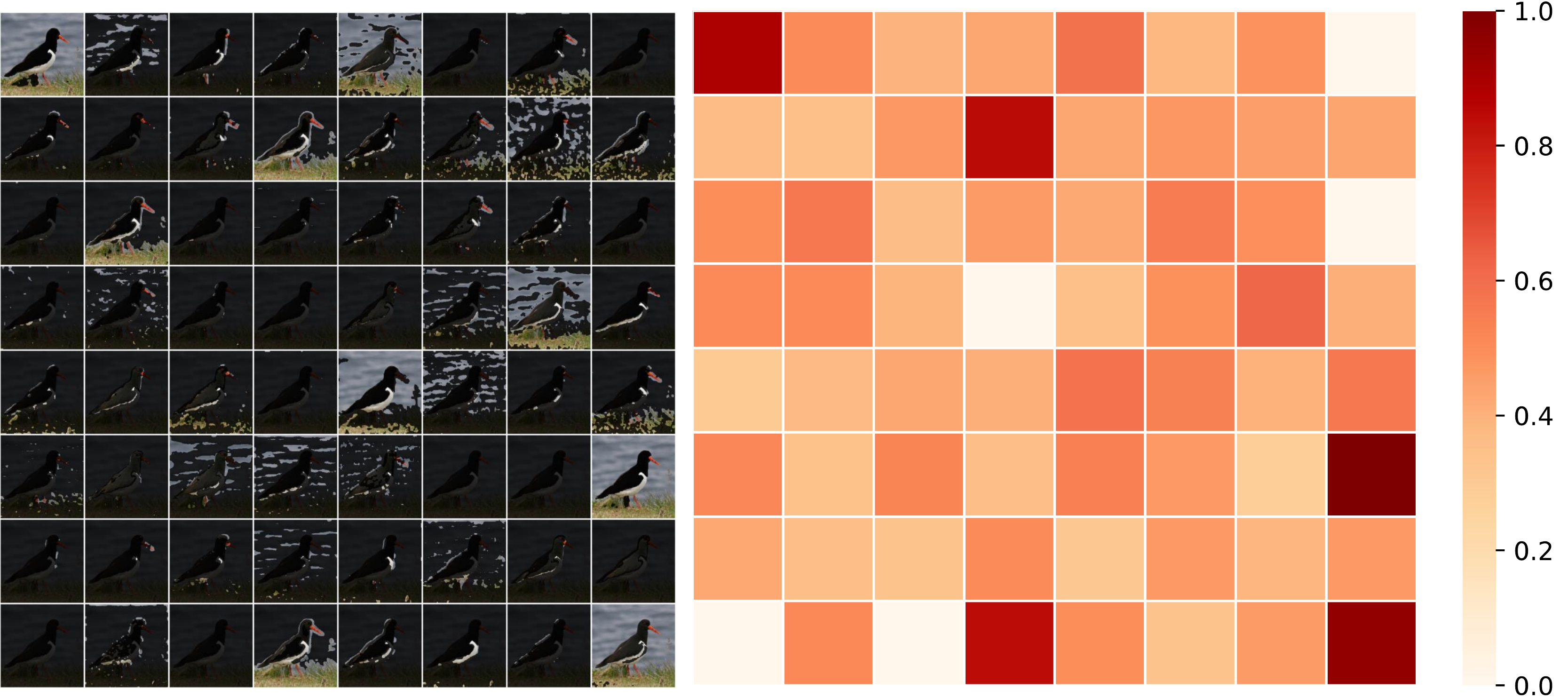}
\end{center}
\vspace{-.5em}
   \caption{Visualization of the input feature maps and their KSE indicator values at the \emph{Block1-Unit1-Conv2} layer of ResNet-50. }
   \vspace{-1em}
\label{fig:vis}
\end{figure}

We visualize the input feature maps and the corresponding KSE indicator values at the \emph{Block1-Unit1-Conv2} layer of ResNet-50 to reveal their connection.
As shown in Fig. \ref{fig:vis}, the input image contains a bird.
When the value of the indicator is smaller, its corresponding feature map provides less information of the bird.
On the contrary, when the value is close to $1$, the feature map has both the bird and the background information.
Therefore, our KSE method can accurately discriminate the feature maps, and effectively judge their importance.

%------------------------------------------------------------------------
\section{Conclusion}
\label{sec:6}

In this paper, we first investigate the problem of CNN compression from a novel \textit{interpretable} perspective and discover that the sparsity and information richness are the key elements to evaluate the importance of the feature maps.
Then we propose kernel sparsity and entropy (KSE) and combine them as an indicator to measure this importance in a \textit{feature-agnostic} manner.
Finally, we employ kernel clustering to reduce the number of kernels based on the KSE indicator and fine-tune the compressed network in a few epochs.
The networks compressed using our approach achieve better results than state-of-the-art methods.
For future work, we will explore a more rigorous theoretical proof with bounds/conditions to prove the relationship between feature map and kernels.
The code available at \url{https://github.com/yuchaoli/KSE}.

\section*{Acknowledgments}
This work is supported by the National Key R\&D Program (No.2017YFC0113000, and No.2016YFB1001503), the Natural Science Foundation of China (No.U1705262, No.61772443, No.61402388 and No.61572410), the Post Doctoral Innovative Talent Support Program under Grant BX201600094, the China Post-Doctoral Science Foundation under Grant 2017M612134, Scientific Research Project of National Language Committee of China (Grant No. YB135-49), and Natural Science Foundation of Fujian Province, China (No. 2017J01125 and No. 2018J01106).

{\small
\bibliographystyle{ieee_fullname}
\bibliography{cvpr_eg}
}

\end{document}